\documentclass[10pt,twocolumn,letterpaper]{article}

\usepackage{cvpr}              
\definecolor{cvprblue}{rgb}{0.21,0.49,0.74}
\usepackage[pagebackref,breaklinks,colorlinks,allcolors=cvprblue]{hyperref}
\usepackage{multirow}

\usepackage[table]{xcolor}
\definecolor{DustyRose}{HTML}{E8B4B8}
\definecolor{Peach}{HTML}{F4D9C6}
\definecolor{SageGreen}{HTML}{D4E4BC}
\definecolor{SkyBlue}{HTML}{B8D4E8}
\definecolor{Lavender}{HTML}{C8B8E8}
\definecolor{Mauve}{HTML}{E8C4D4}
\definecolor{lightgray}{gray}{0.85}
\usepackage{cuted}
\usepackage{arydshln}
\usepackage[most]{tcolorbox}

\def\paperID{xxxx} 
\def\confName{xxxx}
\def\confYear{xxxx}

\title{Omni-Judge: Can Omni-LLMs Serve as Human-Aligned Judges \\ for Text-Conditioned Audio-Video Generation?
}

\author{
Susan Liang{$^1$}, Chao Huang{$^1$}, Filippos Bellos{$^2$}, Yolo Yunlong Tang{$^1$}, \\ Qianxiang Shen{$^1$}, Jing Bi{$^1$}, Luchuan Song{$^1$}, Zeliang Zhang{$^1$}, Jason J Corso{$^2$}, Chenliang Xu{$^1$} \\
{$^1$}University of Rochester\\
{$^2$}University of Michigan, Ann Arbor\\
\small{\url{https://liangsusan-git.github.io/project/omni_judge/}}
}

\begin{document}
\maketitle

\begin{abstract}
State-of-the-art text-to-video generation models such as Sora 2 and Veo 3 can now produce high-fidelity videos with synchronized audio directly from a textual prompt, marking a new milestone in multi-modal generation. However, evaluating such tri-modal outputs remains an unsolved challenge. Human evaluation is reliable but costly and difficult to scale, while traditional automatic metrics, such as FVD, CLAP, and ViCLIP, focus on isolated modality pairs, struggle with complex prompts, and provide limited interpretability.
Omni-modal large language models (omni-LLMs) present a promising alternative: they naturally process audio, video, and text, support rich reasoning, and offer interpretable chain-of-thought feedback. Driven by this, we introduce Omni-Judge, a study assessing whether omni-LLMs can serve as human-aligned judges for text-conditioned audio-video generation.
Across nine perceptual and alignment metrics, Omni-Judge achieves correlation comparable to traditional metrics and excels on semantically demanding tasks such as audio-text alignment, video-text alignment, and audio-video-text coherence. It underperforms on high-FPS perceptual metrics, including video quality and audio-video synchronization, due to limited temporal resolution. Omni-Judge provides interpretable explanations that expose semantic or physical inconsistencies, enabling practical downstream uses such as feedback-based refinement. 
Our findings highlight both the potential and current limitations of omni-LLMs as unified evaluators for multi-modal generation.
\end{abstract}    
\section{Introduction}
\label{sec:intro}
State-of-the-art text-to-video generation models \cite{liu2024syncflow,liu2024tell,low2025ovi,weng2025audio,zhao2025uniform,cheng2024taming,wang2025universe,meta2024moviegen,zhang2024asva,mao2024tavgbench,wang2024avdit,yariv2024audiotovideo,ren2024stav2a,zhang2024foleycrafter,mo2024t2avsync,mo2023diffava,huang2023makeanaudio2,ruan2023mmdiffusion} such as Sora 2 \cite{openai_sora2_2025} and Veo 3 \cite{google_veo3_2025} can now produce high-fidelity videos with synchronized audio directly from a textual prompt, marking a new milestone in multi-modal generation. These models unlock new creative, educational, and industrial applications, but also introduce a pressing need for reliable evaluation of tri-modal outputs that include text, video, and audio. Despite rapid progress in generation, evaluation remains an unsolved challenge.

Human evaluation is widely regarded as the gold standard for judging perceptual quality, semantic correctness, and cross-modal coherence \cite{meta2024moviegen}. Yet it is expensive, slow, and difficult to scale, especially when evaluating hundreds or thousands of audio-video samples. Existing automatic metrics alleviate this burden but have significant limitations. Traditional video metrics such as FVD \cite{unterthiner2019fvd} and CLIP-based scores \cite{radford2021learning, xu2021videoclip} assess visual realism or video-text similarity but largely ignore audio content. Audio-focused metrics such as FAD \cite{kilgour2018fr} and CLAPScore \cite{clap} capture audio quality or audio-text alignment but overlook visual semantics. Audio-visual metrics such as FAVD \cite{hershey2017cnn,tran2015learning}, FATD \cite{mikolov2013efficient}, and FAVTD \cite{mikolov2013efficient,mo2024text}, and Align Acc \cite{luo2023diff} operate at the embedding or distributional level but cannot interpret prompt semantics or provide fine-grained reasoning. Overall, these metrics evaluate each modality or modality pair in isolation, struggle with complex prompts, and produce outputs that are difficult to interpret.

In parallel, recent omni-modal large language models (omni-LLMs) \cite{openai2024gpt4o,deepmind2025gemini25,videollama,videosalmonn,videosalmonn2,qwen25omni,qwen3omni} such as GPT-4o \cite{openai2024gpt4o}, Gemini 2.5 \cite{deepmind2025gemini25}, and Qwen3-Omni \cite{qwen3omni} offer promising capabilities for unified perception and reasoning across text, image, audio, and video. These models jointly encode multiple modalities, demonstrate strong semantic understanding, and often produce interpretable chain-of-thought explanations. This emerging capability raises an open research question: \textbf{can omni-LLMs serve as human-aligned judges for text-conditioned audio-video generation?} Unlike traditional metrics, omni-LLMs have the potential to evaluate all three modalities jointly, reason about semantic correctness and temporal consistency, and provide interpretable explanations. At the same time, their performance on fine-grained perceptual tasks remains unclear, especially given their limited temporal resolution during inference. This motivates a careful, empirical study of using omni-LLMs as a multi-modal generation judge.

In this work, we introduce Omni-Judge, a systematic study of using omni-LLMs to evaluate text-conditioned audio-video generation. We design nine metric-specific instructions, covering visual quality, audio quality, semantic alignment, temporal alignment, and aesthetics, and evaluate Omni-Judge on videos generated by Sora 2 and Veo 3 using prompts curated from VidProM \cite{wang2024vidprom}. We compare Omni-Judge against traditional metrics and human annotations across all dimensions.

Our findings reveal that omni-LLMs offer both strengths and limitations. Omni-Judge achieves correlation comparable to traditional metrics and excels on semantically complex tasks such as audio-text alignment, video-text alignment, and tri-modal coherence. However, it struggles on high-FPS perceptual metrics such as video quality and audio-video synchronization, due to limited temporal resolution. Importantly, Omni-Judge provides natural-language explanations that identify semantic or physical inconsistencies, enabling downstream applications such as feedback-based refinement of generative models. Our contributions are listed as follows:
\begin{itemize}
    \item We present Omni-Judge, a systematic study evaluating whether omni-LLMs can serve as human-aligned judges for text-conditioned audio-video generation, supported by a generation benchmark with human annotations.
    \item Through extensive comparison with traditional metrics, we show that omni-LLMs excel on semantic and reasoning-heavy metrics but underperform on high-FPS perceptual tasks, revealing key strengths and limitations of current omni-models.
    \item We demonstrate that Omni-Judge provides interpretable feedback that can localize errors and support downstream refinement of multi-modal generative models.
\end{itemize}

\section{Related Work}
\label{sec:related_work}
\subsection{Traditional Metrics for Content Generation}
Evaluating generative video models has primarily relied on metrics for visual fidelity, audio quality, and pairwise modality alignment. Frechet Video Distance (FVD) \cite{unterthiner2019fvd} measures realism by comparing distributional statistics of video embeddings, but it is agnostic to the input prompt and insensitive to semantic correctness. CLIP-based metrics \cite{radford2021learning, xu2021videoclip} improve text-video alignment by leveraging contrastive vision-language models, yet they often over-reward object presence and fail to capture temporal consistency or narrative accuracy. For audio evaluation, CLAP \cite{clap} extends contrastive learning to audio-text pairs, while ImageBind \cite{girdhar2023imagebind} provides unified multi-modal embeddings across audio, image, and text. However, these methods are typically used in pairwise fashion and lack prompt-aware grounding. To capture audio-visual correspondence, several works propose audio-visual consistency metrics such as FAVD \cite{hershey2017cnn,tran2015learning}, FATD \cite{mikolov2013efficient}, and FAVTD \cite{mikolov2013efficient,mo2024text}, Align Acc \cite{luo2023diff}. Although effective for detecting modality mismatch, these metrics operate at the distribution level, ignore textual intent, and cannot explain semantic or contextual failures.

\subsection{LLM-Based Evaluation for Video and Images}
The rise of multi-modal large language models (MLLMs) \cite{Qwen-VL,Qwen2-VL,Qwen2.5-VL,chen2024internvl,zhu2025internvl3,liu2023llava,liu2023improvedllava,liu2024llavanext,openai2024gpt4o,deepmind2024gemini15,deepmind2025gemini25,anthropic2024claude3,bytedance2024doubao,team2025kimi,dong2025qianfan} has enabled LLM-based evaluation for visual and video generation. Early work \cite{wu2024gpt} using GPT-4V \cite{openai2023gpt4} demonstrated that prompting vision-language models to score or describe generated content can yield judgments more consistent with human preferences than traditional metrics. Benchmark frameworks such as VBench \cite{huang2024vbench} and EvalCrafter \cite{liu2024evalcrafter} incorporate LLM reasoning as part of a broader suite of sub-metrics, though their evaluations remain modular and treat different aspects independently. More recent efforts aim for holistic LLM-based evaluators. GRADEO \cite{mou2025gradeo} fine-tunes an MLLM on human-written video critiques to produce natural-language assessments and per-dimension scores. Video-Bench \cite{han2025video} explores few-shot prompting and chain-of-query reasoning to improve scoring stability. These studies show that reasoning-based evaluation achieves significantly higher correlation with human judgments than conventional metrics. However, a major limitation is that existing LLM-based evaluators focus almost exclusively on silent videos. With the emergence of audio-video generators such as Sora 2 \cite{openai_sora2_2025} and Veo 3 \cite{google_veo3_2025}, this omission becomes increasingly problematic, motivating unified evaluation methods that can jointly assess audio, video, and text.

\subsection{Omni-Modal LLMs}
Recent omni-modal foundation models \cite{openai2024gpt4o,deepmind2025gemini25,videollama,videosalmonn,videosalmonn2,qwen25omni,qwen3omni} such as Qwen3-Omni \cite{qwen3omni} and Gemini 2.5 \cite{deepmind2025gemini25} extend LLMs to jointly process audio, video, and text. Trained on large-scale tri-modal data, these models can perform end-to-end reasoning over synchronized audio-video inputs and form unified representations across modalities, enabling more coherent understanding of scene semantics and cross-modal dependencies. Motivated by the capabilities of omni-modal models, we study whether an omni-LLM can function as a unified evaluator for text-conditioned audio-video generation. This unified approach offers the potential to capture cross-modal inconsistencies, provide interpretable judgments, and better reflect human perception than traditional pairwise metrics.
\section{Method}
\label{sec:method}
To examine whether omni-LLMs can act as human-aligned judges for audio-video generation, we construct a comprehensive evaluation pipeline. We begin by assembling a diverse prompt set and generating multi-modal outputs with state-of-the-art models. We then gather human ratings as ground truth and design the Omni-Judge framework, which evaluates nine dimensions of perceptual and semantic quality using task-specific instructions.
\begin{figure}[t]
    \centering
    \includegraphics[width=0.8\linewidth]{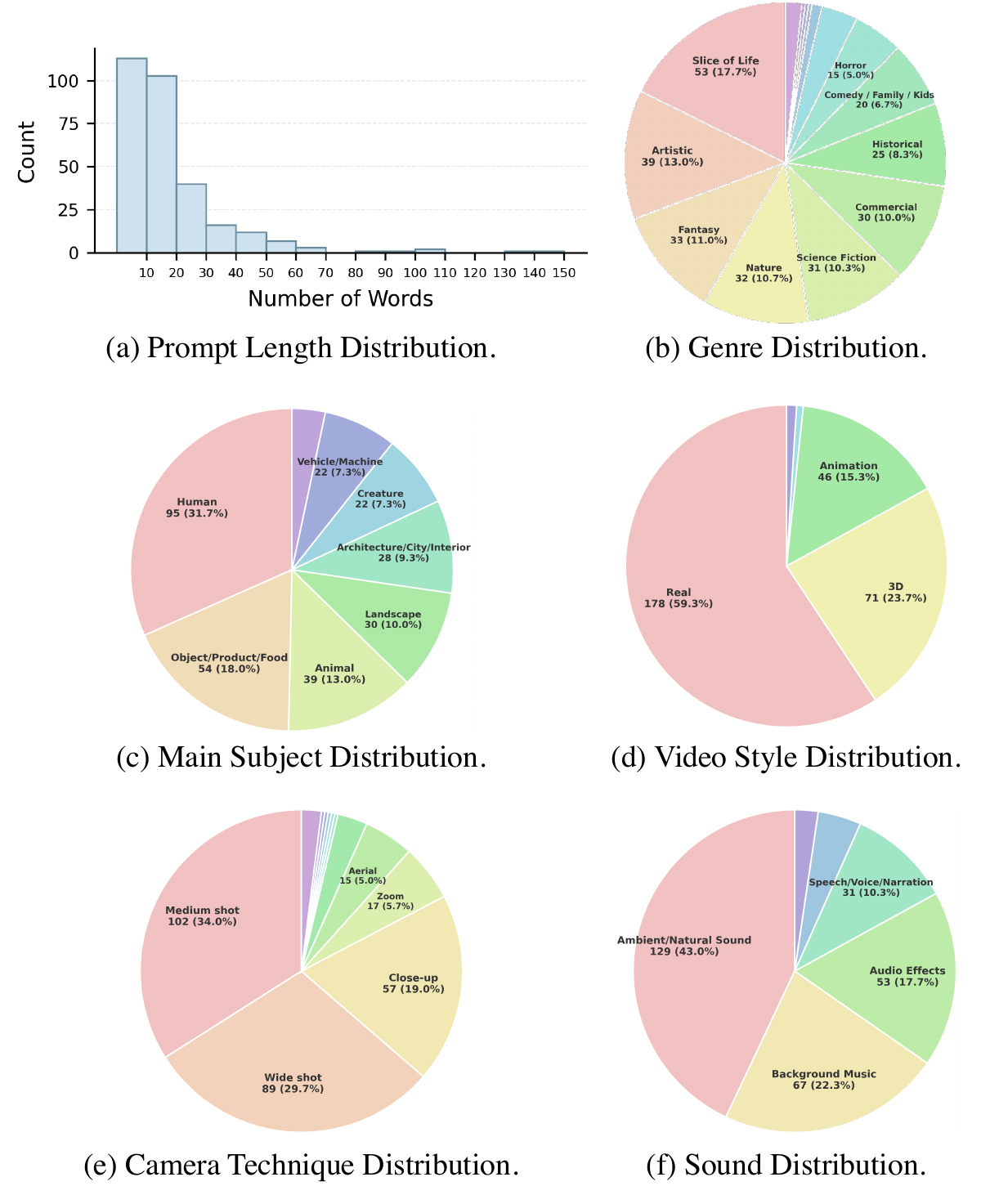}
    \vspace{-2mm}
    \caption{Statistical analysis of the collected 300 prompts from VidProM, illustrating the distributions across linguistic length, film genres, main subjects, visual styles, camera techniques, and sound dimensions.}
    \label{fig:prompt_distribution}
    \vspace{-2mm}
\end{figure}
\subsection{Prompt Collection and Analysis} 
Firstly, we construct a diverse prompt set derived from VidProM \cite{wang2024vidprom}, a large-scale real-user prompt gallery for text-to-video generation. Compared with earlier text-to-image datasets such as DiffusionDB \cite{wang2023diffusiondb}, VidProM better reflects realistic video-generation requests from end users, covering a broad range of topics, visual styles, and sound conditions. From this corpus, we randomly sample 300 representative prompts to serve as evaluation queries for our benchmark.

To better understand the characteristics of our collected prompts, we perform a comprehensive statistical analysis across multiple dimensions, including linguistic complexity, visual style, cinematic composition, and sound specification. \cref{fig:prompt_distribution} summarizes these distributions.

\noindent\textbf{Prompt Length Distribution.}
As shown in Figure~\ref{fig:prompt_distribution} (a) (top-left), most prompts are concise: over 70\% contain fewer than 20 words, while only a few exceed 60 words. This aligns with real user behavior, where short, descriptive sentences are prevalent in generative video systems.

\noindent\textbf{Film Genre Distribution.}
\cref{fig:prompt_distribution} (b) (top-right) illustrates that prompts span multiple cinematic genres. The largest portion belongs to \textit{Slice of Life} (17.7\%), followed by \textit{Artistic}, \textit{Fantasy}, \textit{Science Fiction}, and \textit{Nature}. The presence of \textit{Commercial} and \textit{Historical} scenes suggests that prompts are not restricted to entertainment but also include creative and advertising contexts, providing a broad testbed for generalization of text-to-video models.

\noindent\textbf{Main Subject Categories.}
As shown in \cref{fig:prompt_distribution} (c) (middle-left), human-centric scenes dominate (31.7\%), consistent with the natural focus of storytelling videos. Other frequent subjects include \textit{Objects / Products / Food}, \textit{Animals}, \textit{Architecture / City / Interior}, and \textit{Vehicles or Machines}. Such diversity ensures that diverse subjects and scenarios are covered for audio-video alignment evaluation.

\noindent\textbf{Visual Style.}
In \cref{fig:prompt_distribution} (d) (middle-right), most prompts describe real-world visual styles (59.3\%), while 3D and animation settings also appear frequently. This balance allows us to test model robustness across photorealistic, rendered, and stylized domains.

\noindent\textbf{Camera Technique and Composition.}
As shown in \cref{fig:prompt_distribution} (e) (bottom-left), medium shots (34.0\%) and wide shots (29.7\%) are the most common, followed by close-ups and zoom scenes. This reflects a natural mix of narrative and cinematic compositions, facilitating evaluations of video-text alignment and motion realism.

\noindent\textbf{Sound Dimension.}
Finally, \cref{fig:prompt_distribution} (f) (bottom-right) summarizes the audio aspects explicitly mentioned in prompts. The majority involve ambient or natural sounds (43.0\%), followed by background music, audio effects, and speech or voice narration. This sound diversity enables multi-modal testing over ambience consistency, soundtrack-motion correspondence, and voice-scene synchronization.

\begin{figure*}[t]
    \centering
    \includegraphics[width=\linewidth]{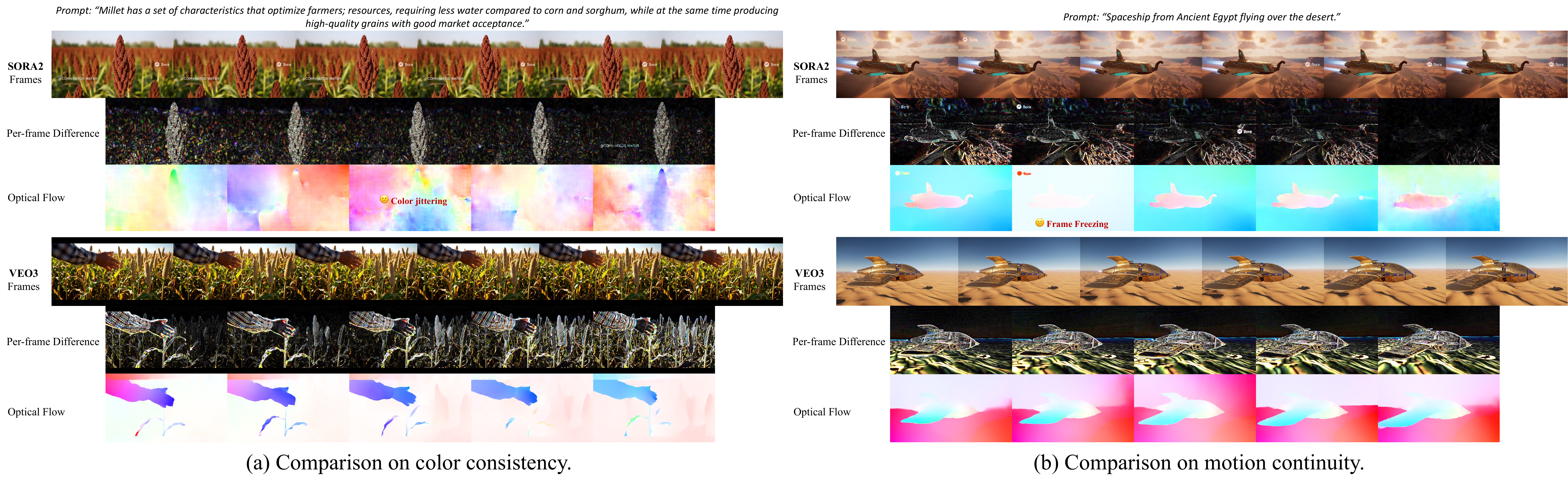}
    \vspace{-5mm}
    \caption{(a) Comparison on color consistency. Sora 2 shows noticeable color jittering and brightness shifts (visible in optical flow), while Veo 3 produces sharper and more stable frames with smoother temporal consistency. (b) Comparison on motion continuity. Sora 2 exhibits frame freezing with static optical flow, whereas Veo 3 maintains fluent motion and higher temporal stability across frames.}
    \vspace{-3mm}
    \label{fig:color_jitter_frame_freeze}
\end{figure*}

\subsection{Video Generation}

To obtain high-quality multi-modal content corresponding to the collected prompts, we utilize two state-of-the-art text-to-video models: Sora 2 by OpenAI~\cite{openai_sora2_2025} and Veo 3 by Google DeepMind~\cite{google_veo3_2025}. Both represent the frontier of generative modeling, capable of synthesizing not only temporally coherent and visually realistic videos but also producing synchronized \textbf{audio tracks} (including ambient sound, music, effects, and dialogue). This unified audio-video generation capability marks a major departure from prior text-to-video systems \cite{singer2022make,yang2024cogvideox,henschel2025streamingt2v}, which typically produced silent clips or required separate sound pipelines.

For each of the 300 prompts curated in our benchmark, we generate corresponding videos using both Sora 2 and Veo 3 to ensure direct cross-model comparability. These outputs enable systematic analysis of visual realism, temporal dynamics, and cross-modal alignment. We present the generation settings in the Appendix. 

\subsection{Human Evaluation Protocol}
To establish a human-grounded benchmark, we conduct a controlled evaluation of all generated videos. Each clip is assessed along four major aspects — quality, semantic alignment, temporal alignment, and aesthetics — covering nine metrics in total. Annotators view the text prompt, watch the audio-video clip, and assign integer scores from 1 to 5 (Very Poor to Very Good). Six Ph.D.\ students with experience in audio-video generation perform all ratings.

The \textbf{quality} aspect measures perceptual fidelity: video quality evaluates clarity, color consistency, and the absence of artifacts such as flickering or distortion, while audio quality assesses naturalness, balance, noise level, and timbral realism. \textbf{Semantic alignment} examines whether modalities convey consistent meaning: audio-video alignment checks whether sounds match visual actions, audio-text alignment evaluates sound relevance to the prompt, video-text alignment considers whether the visuals follow the prompt description, and audio-video-text alignment measures overall tri-modal coherence. \textbf{Temporal alignment} focuses on audio-video synchronization, including whether sound timing matches visual motion such as lip movement or object interactions. Finally, the \textbf{aesthetic} aspect captures stylistic impression, where video aesthetic reflects cinematography and visual appeal, and audio aesthetic considers expressiveness and overall sonic pleasantness. Additional annotation details are provided in the Appendix.

\subsection{Human Preference Analysis}

\begin{figure*}[t]
    \centering
    \includegraphics[width=\linewidth]{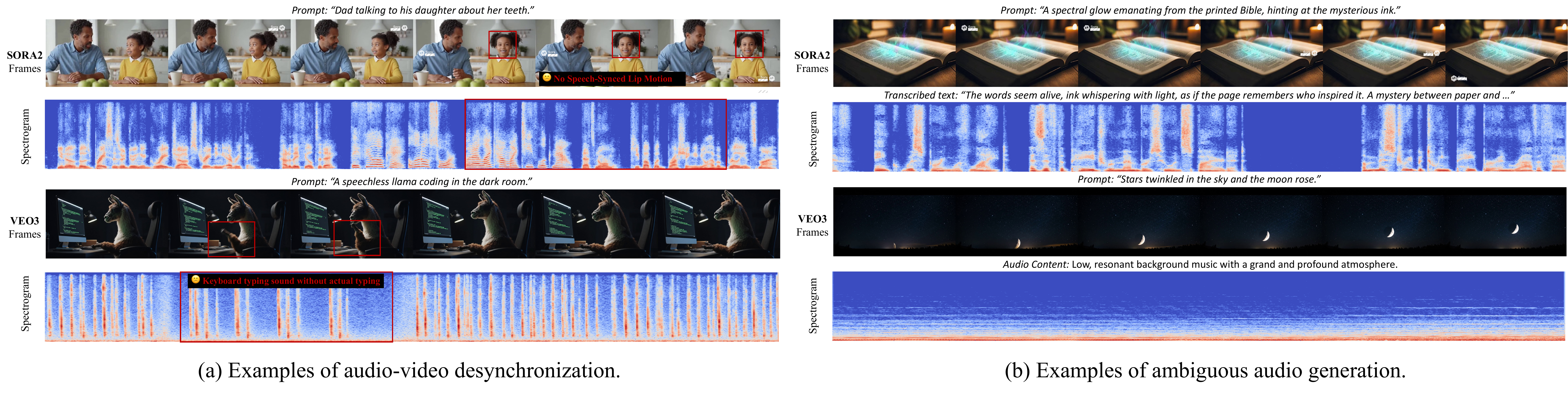}
    \vspace{-7mm}
    \caption{(a) Examples of audio-video desynchronization. Sora 2 produces speech without matching lip motion (top), while Veo 3 generates keyboard typing sounds without corresponding hand movement (bottom), illustrating temporal misalignment between sound and action. (b) Examples of ambiguous audio generation. The models adopt different strategies to match sound with visuals: narration describing the scene (top), and background music expressing emotion and tone (bottom).}
    \vspace{-2mm}
    \label{fig:av_not_sync_audio_ambiguity}
\end{figure*}

\begin{figure}[t]
    \centering
    \includegraphics[width=\linewidth]{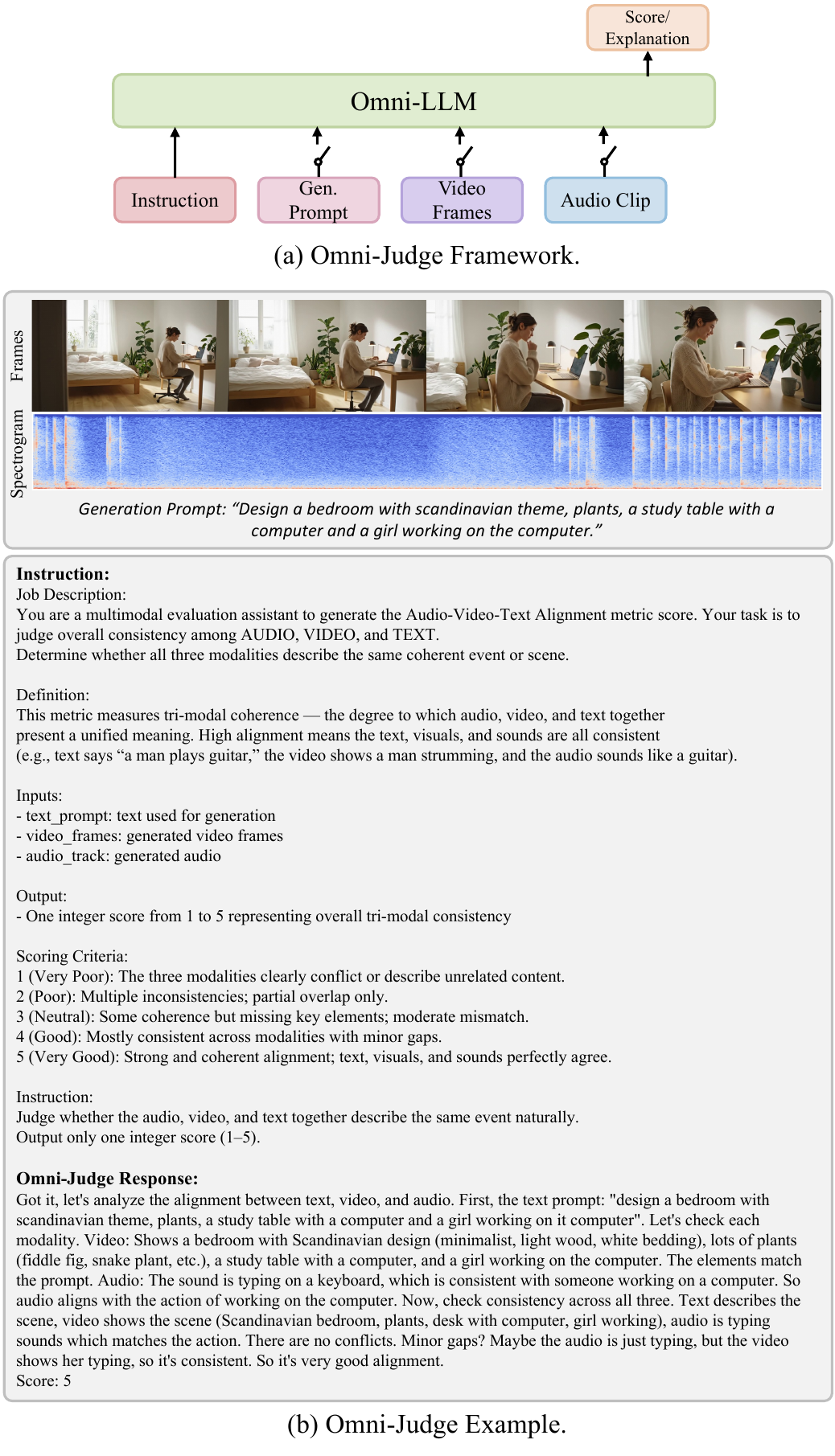}
    \vspace{-7mm}
    \caption{Overview of the Omni-Judge framework.
    (a) Each metric is defined by a task-specific instruction and the required modality inputs. The omni-LLM (Qwen3-Omni) jointly processes text, video, and audio to output a score and explanation. 
    (b) Example of the \textit{audio-video-text alignment} metric, where Omni-Judge reasons step by step across modalities to assess coherence and produce an interpretable score.
    }
    \vspace{-3mm}
    \label{fig:omni_judge_framework}
\end{figure}

\begin{table}[t]
    \centering
    \caption{Human evaluation results averaged over 300 prompts. Scores range from 1 to 5.}
    \vspace{-1mm}
    \label{tab:mos}
    \resizebox{0.4\textwidth}{!}{
    \begin{tabular}{lcc}
    \toprule
        Metrics & Sora 2 \cite{openai_sora2_2025} & Veo 3 \cite{google_veo3_2025} \\
    \hline
        Video Quality & 4.323 & 4.490 \\
        Audio Quality & 4.193 & 3.977 \\
        Audio-Video Alignment & 4.010 & 3.983 \\
        Audio-Text Alignment & 3.817 & 3.770 \\
        Video-Text Alignment & 4.733 & 4.453 \\
        Audio-Video-Text Alignment & 4.020 & 3.923 \\
        Audio-Video Synchronization & 3.410 & 3.600 \\
        Video Aesthetic & 3.813 & 4.220 \\
        Audio Aesthetic & 3.373 & 3.467 \\
    \bottomrule
    \end{tabular}
    }
    \vspace{-6mm}
\end{table}

After collecting all human annotations, we compute the mean opinion scores for all nine metrics (Table~\ref{tab:mos}). The results reveal how participants perceive the strengths and weaknesses of Sora 2 and Veo 3 and provide guidance for designing human-aligned evaluation metrics.

\textit{(1) Video quality and aesthetic reveal that visual precision defines preference.}
Participants consistently preferred videos with fewer perceptual artifacts such as color jittering, frame freezing, and brightness fluctuations. Veo 3 scores higher in both video quality and video aesthetic (4.49 and 4.22) than Sora 2 (4.32 and 3.81), producing sharper and more stable frames. As illustrated in \cref{fig:color_jitter_frame_freeze}, Sora 2 exhibits noticeable color shifts and temporary motion stalls, while Veo 3 maintains smoother temporal consistency. These findings highlight the importance of fine-grained visual fidelity in multi-modal evaluation.

\textit{(2) Audio-video synchronization remains a challenge.}
Both models show limited temporal correspondence between sound and motion, with low synchronization scores (3.41 for Sora 2, 3.60 for Veo 3). Annotators were sensitive to errors such as speech without lip motion or typing sounds without hand movement (\cref{fig:av_not_sync_audio_ambiguity} (a)). Such mismatches noticeably reduce perceived realism, emphasizing the need for metrics that capture accurate multi-modal timing.

\textit{(3) Ambiguity in audio generation arises from uncertain prompt intent.}
Because many prompts lack explicit sound descriptions, models often infer audio to fill semantic gaps. This leads to moderate audio-text and audio-video-text alignment scores. As shown in \cref{fig:av_not_sync_audio_ambiguity} (b), Sora 2 may introduce narrative voiceovers, while Veo 3 frequently adds background music to convey emotional tone. These behaviors indicate that evaluating audio requires contextual understanding rather than simple sound correspondence.

\textit{(4) Audio aesthetics favor musicality and clarity.}
Participants preferred clips with melodious music or clear speech and penalized noisy, flat, or unbalanced audio. Perceived audio aesthetics were strongly influenced by tonal richness and overall expressiveness.

\textit{(5) Video-text alignment highlights strong prompt-following capabilities.}
Both models achieve high video-text alignment (4.73 for Sora 2, 4.45 for Veo 3), reflecting mature prompt adherence in large-scale video generators. 

\subsection{Omni-Judge Framework}
Building on the findings in the human evaluation and the advance of omni-LLMs, we introduce \textbf{Omni-Judge}, an omni-LLM-based evaluation framework. By leveraging an omni-modal foundation model, Omni-Judge can jointly attend to text, video, and audio, reason over their relationships, and produce scores with interpretable feedback. This endows the metric with three advantages: (i) unified access to all relevant modalities, (ii) multi-modal reasoning that can recognize subtle, context-dependent differences, and (iii) interpretable outputs that expose the model’s decision path.

\paragraph{Framework.}
As illustrated in \cref{fig:omni_judge_framework} (a), Omni-Judge evaluates each metric via a task-specific \textit{system instruction} plus the \textit{modality inputs} required by that metric. We design one instruction per metric (nine in total) that defines its goal, inputs, and a 1--5 scoring rubric. The input bundle then includes the modalities needed for that goal. For example, audio-video temporal alignment uses audio+video without text, audio aesthetics uses audio only, and audio-video-text alignment uses audio+video+text. The instruction and inputs are passed to the omni-LLM, which returns a reasoning path as an explanation and a scalar score. 

\paragraph{System Instruction.}
Each Omni-Judge metric is guided by a carefully designed \textit{system instruction} that defines what the omni-LLM should evaluate and how to reason about it. The design of these prompts is directly informed by our human annotation analysis, aiming to align the model’s evaluation criteria with real human preferences and perceptual judgments. Specifically:  
(1) For \textit{video quality} and \textit{video aesthetics}, the instruction emphasizes detecting visual artifacts such as color jittering, brightness shifts, frame freezing, and temporal discontinuities.  
(2) For \textit{audio-video synchronization}, it focuses on speech-lip alignment and the timing consistency between visual actions and corresponding sounds.  
(3) For \textit{audio-related alignment} metrics (audio-text and audio-video-text), the instruction extends beyond surface-level matching to include contextual reasoning about narration, external dialogue, and background music that conveys mood or emotion.  
(4) For \textit{audio aesthetics}, it guides the model to prefer clips with melodious background music or clear, well-articulated speech while penalizing noisy or monotonous audio.  
These targeted instructions encourage Omni-Judge to apply nuanced, human-aligned reasoning tailored to each evaluation dimension.

\paragraph{Judgment Example.}
To make the evaluation process intuitive, \cref{fig:omni_judge_framework} (b) shows the \textit{audio-video-text alignment} metric. The system instruction defines tri-modal coherence (\textit{``do the three modalities describe the same event/scene?''}) and specifies a scale from 1 to 5. The inputs comprise the generation prompt $T$, sampled video frames $V$, and the synchronized audio clip $A$. Omni-Judge’s response demonstrates step-by-step reasoning: it first checks whether $V$ matches $T$ (e.g., Scandinavian bedroom, plants, desk with computer, a girl working), then verifies whether $A$ is consistent with the visual action (e.g., typing sounds align with the girl working on a computer), and finally assesses tri-modal consistency to produce a score and a brief explanation. This explicit reasoning provides interpretability while yielding a more human-aligned judgment.
\section{Experiments}
\label{sec:exp}

\subsection{Experimental Setup}
\noindent\textbf{Compared Metrics.}
We compare Omni-Judge against a comprehensive set of traditional evaluation metrics covering all major perceptual and alignment dimensions following \cite{liu2024syncflow,liu2024tell,low2025ovi,weng2025audio,zhao2025uniform,cheng2024taming}.  
For video quality, we use Fréchet Video Distance (FVD)~\cite{unterthiner2019fvd} to assess visual realism, DINO-V3-based ID Consistency \cite{simeoni2025dinov3,wang2025universe} to evaluate subject identity preservation, and a RAFT-based motion score \cite{teed2020raft,wang2025universe} to measure temporal smoothness and motion coherence.  
For audio quality, we adopt Fréchet Audio Distance (FAD)~\cite{kilgour2018fr} and the Audio Inception Score (IS)~\cite{salimans2016improved}, both estimating distribution-level fidelity of generated audio relative to natural recordings.  
For cross-modal alignment, we employ several embedding-based similarity metrics: ImageBind (IB)~\cite{girdhar2023imagebind} for multi-modal representation alignment, CLAP~\cite{clap} for text-audio consistency, and ViCLIP~\cite{wang2023internvid} for text-video grounding.  
To evaluate audio-video synchronization, we include Synchformer~\cite{iashin2024synchformer}, which measures the temporal correspondence between audio events and visual motion.  
For aesthetic evaluation, we use Aesthetic Predictor V2.5~\cite{discus0434_aesthetic_predictor_v2_5} and MUSIQ~\cite{ke2021musiq} to assess visual appeal, and AudioBox-Aesthetics~\cite{tjandra2025meta} to estimate perceived pleasantness and stylistic quality of generated audio. 

\noindent\textbf{Evaluation Protocol.}
To assess the reliability of Omni-Judge, we compare its scores with human annotations across all evaluation dimensions. For each metric (e.g., video quality, audio-video synchronization, audio-text alignment), both humans and automatic metrics assign a scalar score to every sample, producing two sequences of scores. We measure their agreement using three correlation coefficients: Kendall’s $\tau_b$ \cite{kendall1938tau}, Spearman’s $\rho$ \cite{spearman1904correlation}, and Pearson’s $r$ \cite{pearson1895correlation}. Together, these statistics capture both rank-order and linear consistency, providing a comprehensive view of perceptual alignment.

\noindent\textbf{Implementation Details.}
We implement Omni-Judge using the Qwen3-Omni model \cite{qwen3omni}, which contains 30B parameters with 3B active during inference. We evaluate both the instruction-tuned and reasoning-augmented variants to compare their evaluation consistency and interpretability. 

\subsection{Quantitative Comparison}
\begin{table*}[t]
    \centering
    \caption{Comparison of audio, video, and multi-modal evaluation metrics across Sora 2 \cite{openai_sora2_2025} and Veo 3 \cite{google_veo3_2025}. The table covers traditional objective metrics, Omni-LLM-based scores, and human evaluations across different perceptual and alignment dimensions. AudioBox-Aesthetics \cite{tjandra2025meta} evaluate the audio aesthetics in terms of Content Enjoyment (CE), Content Usefulness (CU), Production Complexity (PC), and Production Quality (PQ). The correlation columns present three coefficients, including Kendall’s $\tau_b$~\cite{kendall1938tau}, Spearman’s rank correlation $\rho$~\cite{spearman1904correlation}, and Pearson’s linear correlation $r$~\cite{pearson1895correlation}. Cells highlighted in \textcolor{Peach}{Peach} indicate the best performance within each metric category. }
    \label{tab:correlation}
    \resizebox{0.9\textwidth}{!}{
    \begin{tabular}{llcccc}
    \toprule
    \multirow{2}{*}{Category}                       & \multirow{2}{*}{Metric} & \multicolumn{2}{c}{Sora 2 \cite{openai_sora2_2025}}  & \multicolumn{2}{c}{Veo 3 \cite{google_veo3_2025}}   \\
    & & Score & Correlation & Score & Correlation \\
    \hline
    \multirow{3}{*}{Audio-Video Alignment} 
     & IB \cite{girdhar2023imagebind} & 0.227 & \cellcolor{Peach} 0.161/0.209/0.239 & 0.270 & \cellcolor{Peach} 0.241/0.316/0.348 \\
     & Omni-Judge (Ours) & 4.180 & 0.110/0.121/0.201 & 4.123 & 0.146/0.160/0.146 \\
     & Human & 4.010 & - & 3.983 & -\\
    \hdashline
    \multirow{4}{*}{Audio-Text Alignment} 
     & CLAP \cite{clap} & 0.068 & 0.122/0.161/0.185 & 0.136 & 0.171/0.220/0.176 \\
     & IB \cite{girdhar2023imagebind} & 0.151 & 0.124/0.165/0.158 & 0.168 & \cellcolor{Peach} 0.229/0.294/0.280 \\
     & Omni-Judge (Ours) & 2.517 & \cellcolor{Peach} 0.292/0.345/0.282 & 2.183 &  0.224/0.259/0.220 \\
     & Human & 3.817 & - & 3.770 & - \\
    \hdashline
    \multirow{4}{*}{Video-Text Alignment} 
     & ViCLIP \cite{wang2023internvid} & 0.257 & 0.073/0.091/0.062 & 0.256 & \cellcolor{Peach} 0.086/0.107/0.089 \\
     & IB \cite{girdhar2023imagebind} & 0.346 & 0.044/0.056/0.064 & 0.335 & 0.061/0.077/0.040 \\
     & Omni-Judge (Ours) & 4.817 & \cellcolor{Peach} 0.142/0.148/0.154 & 4.720 & 0.074/0.079/0.127 \\
     & Human & 4.733 & - & 4.453 & - \\
    \hdashline
    \multirow{3}{*}{Audio-Video-Text Alignment} 
     & IB \cite{girdhar2023imagebind} & 0.241 & 0.070/0.090/0.093 & 0.258 & \cellcolor{Peach} 0.216/0.280/0.277\\
     & Omni-Judge (Ours) & 4.727 & \cellcolor{Peach} 0.139/0.151/0.136 & 4.570 & 0.203/0.223/0.217\\
     & Human & 4.020 & - & 3.923 & - \\
    \hline
    \multirow{3}{*}{Audio-Video Synchronization} 
     & Synchformer \cite{iashin2024synchformer} & 0.167 & \cellcolor{Peach} 0.178/0.220/0.248 & 0.191 & \cellcolor{Peach} 0.224/0.286/0.298 \\
     & Omni-Judge (Ours) & 4.367 & 0.142/0.156/0.167 & 4.270 & 0.045/0.050/0.136 \\
     & Human & 3.410 & - & 3.600 & - \\   
    \hline
    \multirow{5}{*}{Video Quality} 
     & FVD \cite{unterthiner2019fvd}   & 1075.771 & - & 1038.972 & - \\
     & ID Consistency \cite{simeoni2025dinov3,wang2025universe} & 0.796 & \cellcolor{Peach} 0.064/0.081/0.069 & 0.849 & -0.001/-0.001/-0.010\\
     & Motion Score \cite{teed2020raft,wang2025universe}   & 0.364 & -0.157/-0.201/-0.153 & 0.807 & \cellcolor{Peach} 0.070/0.087/0.081\\
     & Omni-Judge (Ours)   &  4.473     & 0.020/0.021/-0.002       & 4.530 & -0.041/-0.042/-0.031     \\
     & Human  & 4.323      & - &   4.490 & - \\
    \hdashline
    \multirow{4}{*}{Audio Quality}                     
     & FAD \cite{kilgour2018fr}   & 11.261   & - & 7.545 & - \\
     & IS \cite{salimans2016improved}   & 1.010      & - & 1.012 & - \\
     & Omni-Judge (Ours)   & 3.937 & \cellcolor{Peach} 0.161/0.181/0.206 & 3.820 & \cellcolor{Peach} 0.090/0.103/0.123 \\
     & Human  & 4.193      & - &    3.977 & - \\
    \hline
    \multirow{4}{*}{Video Aesthetic} 
     & Aesthetic Predictor V2.5 \cite{discus0434_aesthetic_predictor_v2_5} & 0.421 & \cellcolor{Peach} 0.271/0.351/0.359 & 0.526 & 0.033/0.042/0.069 \\
     & MUSIQ \cite{ke2021musiq} & 0.567 & 0.128/0.171/0.172 & 0.594 & -0.022/-0.028/-0.010 \\
     & Omni-Judge (Ours) & 4.627 & 0.200/0.219/0.298 & 4.643 & \cellcolor{Peach} 0.083/0.089/0.216 \\
     & Human & 3.813 & - & 4.220 & - \\    
    \hdashline
    \multirow{6}{*}{Audio Aesthetic} 
     & AudioBox-Aesthetics \cite{tjandra2025meta} (CE) & 4.504 & 0.209/0.266/0.287 & 4.337 & 0.203/0.264/0.289 \\
     & AudioBox-Aesthetics \cite{tjandra2025meta} (CU) & 5.567 & 0.147/0.189/0.215 & 6.503 & 0.103/0.137/0.163 \\
     & AudioBox-Aesthetics \cite{tjandra2025meta} (PC) & 4.399 & 0.233/0.306/0.308 & 3.238 & \cellcolor{Peach} 0.253/0.324/0.280 \\
     & AudioBox-Aesthetics \cite{tjandra2025meta} (PQ) & 6.026 & 0.063/0.081/0.098 & 6.822 & 0.063/0.080/0.119 \\
     & Omni-Judge (Ours) & 3.457 & \cellcolor{Peach} 0.253/0.285/0.290 & 2.870 & 0.129/0.149/0.162 \\
     & Human & 3.373 & - & 3.467 & - \\
    \bottomrule
    \end{tabular}
    \vspace{-3mm}
    }
\end{table*}

We evaluate how well Omni-Judge and all baseline metrics correlate with human annotations across the nine evaluation dimensions. As reported in \cref{tab:correlation}, Omni-Judge achieves competitive and often strong correlation with human ratings. For many dimensions, its correlations are comparable to or exceed those of traditional metrics. For example, on Sora 2, Omni-Judge attains Kendall’s~$\tau_b$ / Spearman’s~$\rho$ correlations of 0.292/0.345 for audio-text alignment and 0.139/0.151 for audio-video-text alignment, notably outperforming CLAP \cite{clap}, ImageBind \cite{girdhar2023imagebind}, and ViCLIP \cite{wang2023internvid}. Similar trends hold for Veo 3, where Omni-Judge achieves 0.224/0.259 on audio-text and 0.203/0.223 on audio-video-text alignment. These tasks require deeper semantic understanding of the user prompt and cross-modal reasoning capabilities that embedding-based metrics struggle to capture, but omni-LLMs handle naturally. We also note that metrics such as FVD, FAD, and IS do not appear in the correlation columns because they operate over sample distributions rather than producing per-sample scores, making direct correlation with human ratings infeasible.

However, Omni-Judge underperforms on low-level perceptual metrics, particularly video quality and audio-video synchronization. For Sora 2, correlations for video quality (e.g., $\tau_b = 0.020$) and synchronization (e.g., $\tau_b = 0.142$) lag behind baselines such as ID consistency \cite{simeoni2025dinov3,wang2025universe}, RAFT-based motion score \cite{teed2020raft,wang2025universe}, and Synchformer \cite{iashin2024synchformer}. Similar patterns appear for Veo 3. We attribute this issue to the limited temporal resolution of Omni-Judge, which fails to detect subtle visual artifacts such as color jittering, brightness shifts, frame freezing, or fine sound-motion misalignment. These effects require dense temporal sampling, which per-frame evaluation metrics capture more directly. We further investigate this temporal limitation in our ablation study (see \cref{sec:ablation}).

\subsection{Ablation Study}
\label{sec:ablation}
\begin{figure}[t]
    \centering
    \includegraphics[width=\linewidth]{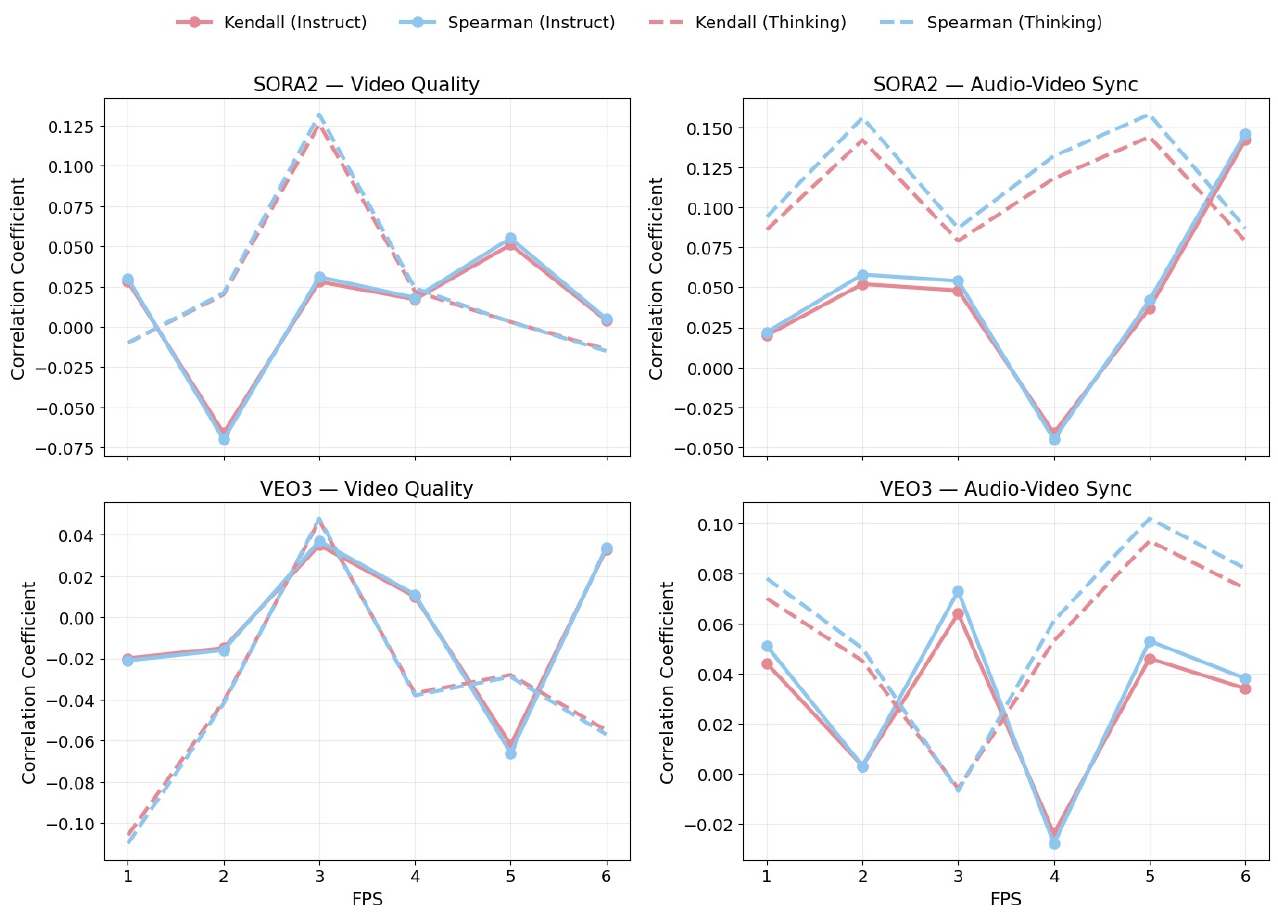}
    \vspace{-3mm}
    \caption{Ablation on model variants and different FPS values. We compare instruction and reasoning models across different frame rates for video quality and audio-video synchronization. Reasoning models perform better overall, and FPS affects performance without a strictly monotonic trend.}
    \vspace{-4mm}
    \label{fig:ablation}
\end{figure}

To examine how input frame rate and model type affect Omni-Judge, we conduct an ablation study on two challenging dimensions: \textit{video quality} and \textit{audio-video temporal alignment}. These metrics require particularly fine temporal sensitivity. Video quality depends on detecting subtle issues such as color shifts and frame discontinuities, while audio–video synchronization requires identifying short-term correspondence between motion and sound. Both tasks therefore benefit from high-FPS visual input.

\paragraph{Instruction vs.\ Reasoning Models.}
Across most settings (\cref{fig:ablation}), the reasoning-oriented model outperforms the instruction-following variant.
For video quality on Sora 2, the reasoning model reaches its strongest correlation at 3\,FPS, with Kendall’s~$\tau_b = 0.126$ and Spearman’s~$\rho = 0.132$, substantially higher than the instruction model at the same FPS ($\tau_b = 0.028$, $\rho = 0.031$).  
For audio-video synchronization on Sora 2, the reasoning model again achieves high correlations, peaking at 5\,FPS with $\tau_b = 0.144$ and $\rho = 0.158$.  
Similar trends appear for Veo 3: the reasoning model reaches $\tau_b = 0.047$ and $\rho = 0.048$ at 3\,FPS for video quality, and up to $\tau_b = 0.093$ and $\rho = 0.102$ at 5\,FPS for audio–video synchronization.  
These results suggest that explicit reasoning helps the model analyze multi-frame patterns more effectively.

\paragraph{Impact of FPS.}
Frame rate also influences performance, but the relationship is not strictly monotonic (\cref{fig:ablation}).  
For video quality, the reasoning model performs best at 3\,FPS for both Sora 2 and Veo 3.  
For audio-video synchronization, the best results appear at 2--5\,FPS depending on the dataset, with Sora 2 peaking at 2\,FPS and Veo 3 peaking at 5\,FPS.  
The instruction model generally benefits from higher FPS, but its improvements are smaller and less stable.

We hypothesize two factors underlying this irregular trend:  
(1) Qwen3-Omni \cite{qwen3omni,qwen25omni} is trained with low frame rates to reduce vision-token costs. These models may not fully utilize high-FPS inputs to recognize subtle difference between consecutive frames.  
(2) Video quality and audio-video synchronization metrics inherently \textit{require} high temporal resolution to identify anomalies such as brief color jittering, small brightness fluctuations, or slight lip-speech misalignment. This creates a mismatch between the training regime of existing omni-models and the temporal precision demanded by these evaluation tasks.

\subsection{Application}
\begin{figure}
    \centering
    \includegraphics[width=\linewidth]{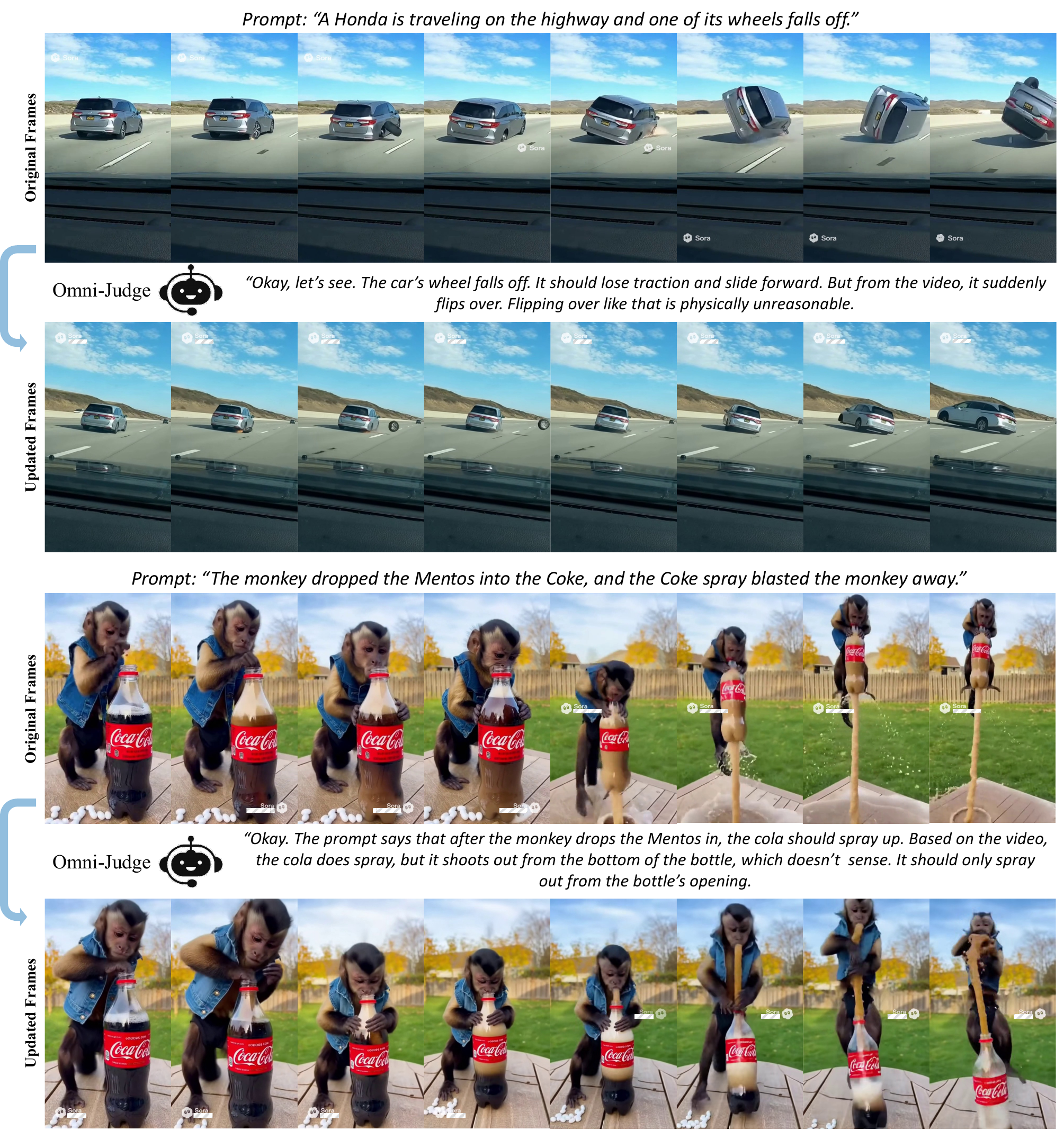}
    \caption{Using Omni-Judge for generation refinement. Given a video and its prompt, Omni-Judge provides textual feedback identifying physically or semantically implausible content. The highlighted issues (car flipping unnaturally; cola spraying from the wrong location) guide the correction of the original frames, producing more plausible updated generations.}
    \vspace{-4mm}
    \label{fig:application}
\end{figure}
Beyond quantitative scoring, Omni-Judge provides natural-language explanations that can be used to identify specific errors in audio-video generations. This interpretability enables a practical application: using Omni-Judge as a feedback module to guide iterative refinement of model outputs. Given a video and its prompt, Omni-Judge can highlight mismatches in physical dynamics, semantic inconsistencies, or implausible visual effects, and these insights can be used to update the generation. As illustrated in \cref{fig:application}, Omni-Judge correctly identifies that a car should lose traction rather than flip abruptly when its wheel detaches, and that cola sprayed by a Mentos reaction should emerge from the bottle’s opening rather than the bottom. Incorporating such feedback allows creators or downstream systems to modify the generation process and achieve improved visual plausibility and prompt adherence.
\section{Conclusion}
\label{sec:conclusion}
We presented Omni-Judge, a systematic study of whether an omni-LLM can serve as a human-aligned judge for text-conditioned audio-video generation. Using Qwen3-Omni, we evaluate nine dimensions of perceptual and semantic quality and compare its scores with human ratings and traditional metrics. Omni-Judge shows strong potential: it matches or exceeds established metrics on semantically complex tasks such as audio-text and tri-modal alignment, while providing interpretable reasoning. However, its limitations on low-level temporal metrics highlight the need for omni-models with stronger high-frame-rate sensitivity. Overall, our findings show both the promise and the current boundaries of using omni-LLMs as unified evaluators for multi-modal generation.

\newpage
{
    \small
    \bibliographystyle{ieeenat_fullname}
    \bibliography{main}
}

\clearpage
\setcounter{page}{1}
\maketitlesupplementary
\appendix

\section{Audio-Video Generation Setting}
The generation settings for Sora 2 \cite{openai_sora2_2025} and Veo 3 \cite{google_veo3_2025} are standardized as follows:
\begin{itemize}
    \item Sora 2: We generate 10-second videos using the \texttt{landscape} aspect ratio. This configuration provides high spatial resolution and cinematic framing suitable for scene-centric content. The generated audio is stereo at a default sampling rate of 96{,}000~Hz.
    \item Veo 3: We generate 8-second videos under the same \texttt{landscape} setting. The slightly shorter duration is consistent with Veo’s default configuration. The generated audio is monophonic at a default sampling rate of 44{,}100~Hz.
\end{itemize}

By leveraging the native audio-video generation capabilities of Sora 2 and Veo 3, our dataset supports comprehensive benchmarking of multi-modal generative systems, advancing evaluation beyond visual fidelity to include auditory quality and audio-visual coherence.

\section{Human Annotation Interface}
\begin{figure*}[t]
    \centering
    \includegraphics[width=\linewidth]{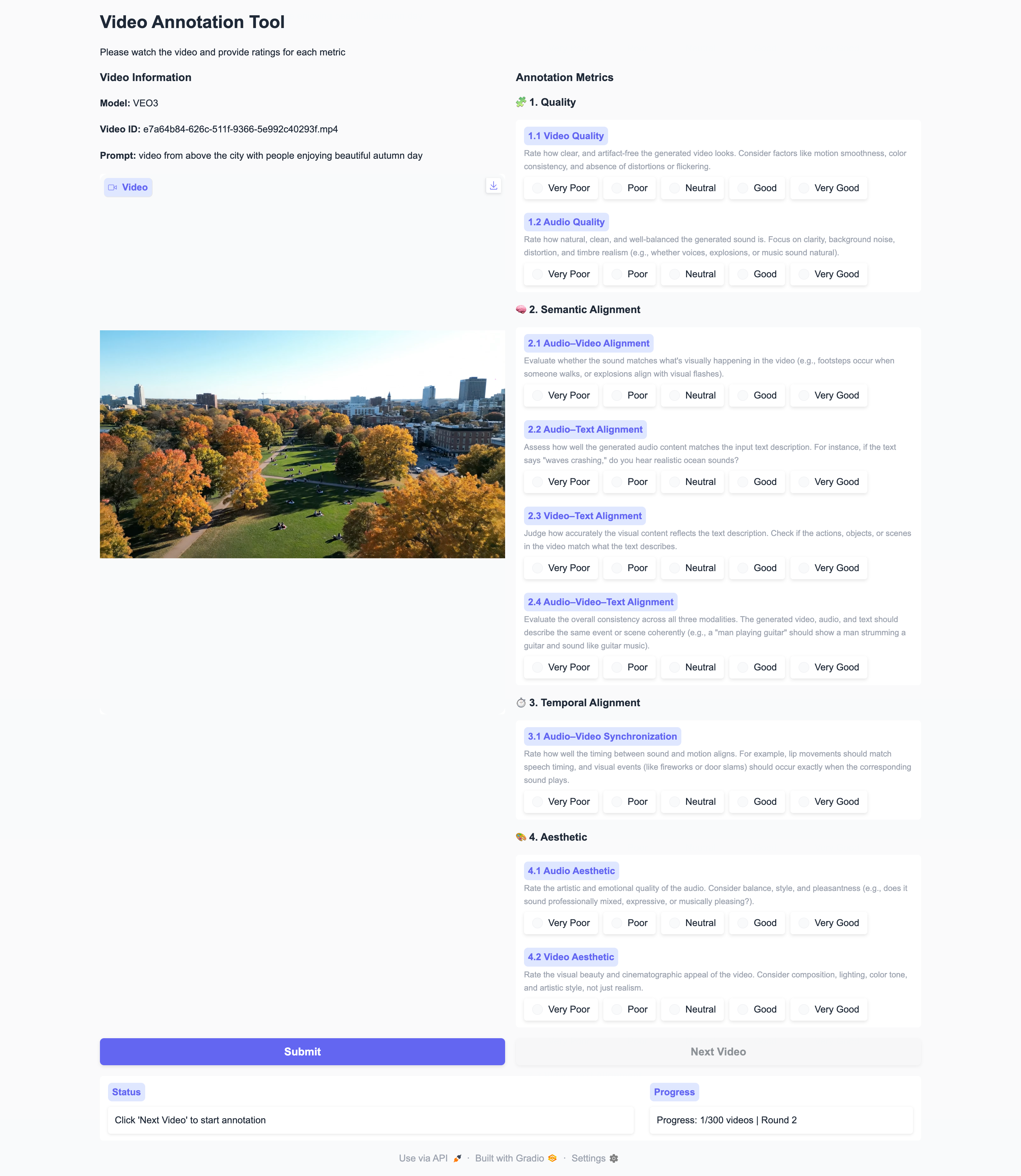}
    \caption{User interface of the Video Annotation Tool. The web-based platform displays the prompt, video with synchronized audio, and rating panels for all nine evaluation metrics. Annotators assess the multi-modal quality, semantic alignment, temporal coherence, and aesthetics.}
    \label{fig:interface}
\end{figure*}
To facilitate consistent human evaluation, we implement a web-based Video Annotation Tool (see \cref{fig:interface}) that integrates video playback, prompt display, and rating panels for all nine metrics. For each video, the annotator is shown the textual prompt and the generated clip with synchronized audio. Annotators can freely replay the video multiple times before submitting their ratings. The collected scores form the basis for quantitative analyses of multi-modal quality, semantic consistency, and aesthetic appeal.

\section{Omni-Judge Instructions}
To guide Omni-Judge toward consistent and interpretable scoring, we design nine metric-specific instruction prompts that correspond directly to the human evaluation rubric. Each prompt explicitly defines the aspect to be judged, covering video quality (\cref{prompt:video_quality}), audio quality (\cref{prompt:audio_quality}), audio-video alignment (\cref{prompt:audio_video_alignment}), audio-text alignment (\cref{prompt:audio_text_alignment}), video-text alignment (\cref{prompt:video_text_alignment}), audio-video-text coherence (\cref{prompt:audio_video_text_alignment}), audio-video synchronization (\cref{prompt:audio_video_temporal_alignment}), audio aesthetic (\cref{prompt:audio_aesthetic}), and video aesthetic (\cref{prompt:video_aesthetic}). and instructs the model to output a single integer score from 1 to 5. These targeted prompts ensure that Omni-Judge reasons about each dimension comprehensively and maintaining alignment with the human annotation protocol.

\begin{figure*}[t]
\centering
\begin{tcolorbox}[colback=black!5!white,colframe=black!75!black,title=Video Quality Prompt]
Job Description: \\
You are a multimodal evaluation assistant designed to generate a Video Quality metric score. Your task is to evaluate the VIDEO QUALITY of a generated clip.\\

Focus only on the visual quality — how stable, clear, and aesthetically pleasing the video looks. \\
Do NOT judge whether the video content is realistic or real-world; high-quality animations, 3D renderings, or stylized videos can also receive high scores.\\
Ignore any aspects of audio or text.\\

Definition:\\
Video Quality measures the overall visual fidelity and consistency of a video. It captures aspects such as sharpness, motion smoothness, temporal stability, color consistency, and the absence of visual artifacts (e.g., blur, flicker, distortion, or broken frames). \\
It focuses purely on how good the video looks, regardless of its style or subject matter.\\

Inputs:\\
- video\_frames: ordered frames from the generated video

Output:\\
- One integer score from 1 to 5 representing the overall visual quality

Scoring Criteria:\\
1 (Very Poor): Severely degraded visuals; extremely low resolution or corrupted frames; heavy flicker or tearing; unwatchable.\\
2 (Poor): Noticeable artifacts, blur, color instability, or motion jitter; distracting or unpleasant to watch.\\
3 (Neutral): Acceptable clarity and smoothness; some compression or blur; generally stable but visually average.\\
4 (Good): Clear, stable visuals with good sharpness, smooth motion, and consistent colors; few minor artifacts.\\
5 (Very Good): Outstanding clarity, detail, and smooth motion; visually appealing and artifact-free across frames.\\

Instruction:\\
Watch the video frames carefully and output a single integer score (1–5) reflecting the overall visual quality.\\
Do not evaluate content realism or meaning — only the perceptual quality of the visuals.
\end{tcolorbox}
\caption{Instruction prompt used by Omni-Judge for the video quality metric.}
\label{prompt:video_quality}
\end{figure*}

\begin{figure*}[t]
\centering
\begin{tcolorbox}[colback=black!5!white,colframe=black!75!black,title=Audio Quality Prompt]
Job Description:\\
You are a multimodal evaluation assistant to generate the Audio Quality metric score. Your task is to evaluate the AUDIO QUALITY of a generated sound clip.\\
Focus purely on how clean, natural, and well-produced the audio sounds. Do not consider alignment or text relevance.\\

Definition:\\
Audio Quality measures technical sound fidelity — clarity, noise level, dynamic range, timbre, and absence
of distortion or clipping. It evaluates whether the sound is pleasant and natural to listen to.\\

Inputs:\\
- audio\_track: generated audio\\

Output:\\
- One integer score from 1 to 5 representing the perceived audio quality\\

Scoring Criteria:\\
1 (Very Poor): Heavily distorted, clipped, or noisy; difficult to understand or unlistenable.\\
2 (Poor): Noticeable noise, hiss, or unnatural timbre; uneven loudness or mixing.\\
3 (Neutral): Acceptable quality; minor distortion or flat sound; average clarity.\\
4 (Good): Clean and balanced; minor imperfections only.\\
5 (Very Good): Studio-level clarity, rich dynamics, and highly natural timbre.\\

Instruction:\\
Listen to the audio carefully and output one integer score (1–5) that best reflects its sound quality.\\
\end{tcolorbox}
\caption{Instruction prompt used by Omni-Judge for the audio quality metric.}
\label{prompt:audio_quality}
\end{figure*}

\begin{figure*}[t]
\centering
\begin{tcolorbox}[colback=black!5!white,colframe=black!75!black,title=Audio-Video Semantic Alignment Prompt]
Job Description:\\
You are an evaluation assistant to generate the Audio–Video Alignment metric score. Your task is to assess how well the AUDIO content semantically and contextually aligns with the VIDEO content — that is, whether the sounds meaningfully fit the visual scene.\\

Definition:\\
Audio–Video Alignment measures the semantic and contextual consistency between sound and visuals.\\
It focuses on whether the audio corresponds to what is happening in the video (e.g., footsteps align with walking, explosion sounds match visual blasts), as well as whether any implicit audio elements support the visual context or emotion.\\
This includes:\\
- Speech or narration that describes or relates to the visual content.\\
- Dialogues that imply or fit the visible events or characters.\\
- Ambient or environmental sounds that match the video setting.\\
- Background music that conveys the same mood, tone, or emotion as the visuals.\\

If the video intentionally contains no sound (e.g., a silent scene where silence is natural or meaningful), assign a neutral score (3) rather than a poor score.\\

Inputs:\\
- video\_frames: generated video frames\\
- audio\_track: generated audio\\

Output:\\
- One integer score from 1 to 5 indicating the semantic and contextual alignment between audio and visuals.\\

Scoring Criteria:\\
1 (Very Poor): Sounds contradict or clash with the visuals; no apparent connection.\\
2 (Poor): Frequent mismatches or unrelated sounds; weak or confusing correspondence.\\
3 (Neutral): Partial alignment; some sounds fit while others are missing or unclear. Also use this when silence is contextually appropriate.\\
4 (Good): Mostly correct and coherent sounds for the visible actions, scenes, or mood.\\
5 (Very Good): Every sound fits naturally; speech, effects, and background audio all enhance the visual narrative and atmosphere.\\

Instruction:\\
Compare the video and audio content carefully. Consider both explicit (e.g., matching actions) and implicit (e.g., narration, background music, ambient fit) alignment. \\
You may internally use speech recognition to interpret spoken content and assess whether it aligns with the visuals, and recognize music or ambient sound types to determine their emotional or contextual fit.\\
Output only one integer score (1–5) representing overall audio–video alignment quality.\\
\end{tcolorbox}
\caption{Instruction prompt used by Omni-Judge for the audio-video semantic alignment metric.}
\label{prompt:audio_video_alignment}
\end{figure*}

\begin{figure*}[t]
\centering
\begin{tcolorbox}[colback=black!5!white,colframe=black!75!black,title=Audio-Text Semantic Alignment Prompt]
Job Description:\\
You are an evaluation assistant to generate the Audio-Text Alignment metric score. Your task is to evaluate how well the AUDIO matches the TEXT PROMPT that was used for generation.\\

Definition:\\
Audio–Text Alignment measures whether the audio correctly represents the events, environment, or mood described by the text prompt (e.g., "a thunderstorm" should have rain and thunder sounds). \\

Inputs:\\
- text\_prompt: the text used for generation \\
- audio\_track: generated audio\\

Output:\\
- One integer score from 1 to 5 indicating semantic alignment between audio and text\\

Scoring Criteria:\\
1 (Very Poor): Audio completely contradicts the text or is irrelevant.\\
2 (Poor): Only loosely related; missing key sounds or context described in the text.\\
3 (Neutral): Some matching features but incomplete or overly generic.\\
4 (Good): Largely consistent with text; only minor omissions or mismatches.\\
5 (Very Good): Perfectly aligned with text; conveys all important details and atmosphere.\\

Instruction:\\
Compare the audio with the text description. Judge how accurately the audio reflects the meaning of the text.
Output one integer score (1–5).
\end{tcolorbox}
\caption{Instruction prompt used by Omni-Judge for the audio-text semantic alignment metric.}
\label{prompt:audio_text_alignment}
\end{figure*}

\begin{figure*}[t]
\centering
\begin{tcolorbox}[colback=black!5!white,colframe=black!75!black,title=Video-Text Semantic Alignment Prompt]
Job Description:\\
You are an evaluation assistant to generate the Video-Text Alignment metric score. Your task is to evaluate how accurately the VIDEO content matches the TEXT PROMPT.\\
Focus on what is shown visually, not how good it looks.\\

Definition:\\
Video–Text Alignment measures semantic consistency between the visual content and the textual description.\\
It evaluates whether the video correctly depicts the objects, actions, or scenes described by the text.\\

Inputs:\\
- text\_prompt: text used for generation\\
- video\_frames: generated video frames\\

Output:\\
- One integer score from 1 to 5 indicating how well the video matches the text\\

Scoring Criteria:\\
1 (Very Poor): The video contradicts the text or is completely unrelated.\\
2 (Poor): Some vague connection but key content is missing or incorrect.\\
3 (Neutral): Partial match; some correct elements but incomplete representation.\\
4 (Good): Mostly faithful to the text with small inaccuracies or missing details.\\
5 (Very Good): Strong, detailed visual match to the text; captures the described scene perfectly.\\

Instruction\\
Compare the text prompt with what you see in the video frames and output one integer score (1–5).
\end{tcolorbox}
\caption{Instruction prompt used by Omni-Judge for the video-text semantic alignment metric.}
\label{prompt:video_text_alignment}
\end{figure*}

\begin{figure*}[t]
\centering
\begin{tcolorbox}[colback=black!5!white,colframe=black!75!black,title=Audio-Video-Text Semantic Alignment Prompt]
Job Description:\\
You are a multimodal evaluation assistant to generate the Audio-Video-Text Alignment metric score. Your task is to judge overall consistency among AUDIO, VIDEO, and TEXT.\\
Determine whether all three modalities describe the same coherent event or scene.\\

Definition:\\
This metric measures tri-modal coherence — the degree to which audio, video, and text together present a unified meaning. High alignment means the text, visuals, and sounds are all consistent (e.g., text says “a man plays guitar,” the video shows a man strumming, and the audio sounds like a guitar).\\

Inputs:\\
- text\_prompt: text used for generation\\
- video\_frames: generated video frames\\
- audio\_track: generated audio\\

Output:\\
- One integer score from 1 to 5 representing overall tri-modal consistency\\

Scoring Criteria:\\
1 (Very Poor): The three modalities clearly conflict or describe unrelated content.\\
2 (Poor): Multiple inconsistencies; partial overlap only.\\
3 (Neutral): Some coherence but missing key elements; moderate mismatch.\\
4 (Good): Mostly consistent across modalities with minor gaps.\\
5 (Very Good): Strong and coherent alignment; text, visuals, and sounds perfectly agree.\\

Instruction:\\
Judge whether the audio, video, and text together describe the same event naturally.\\
Output only one integer score (1–5).\\
\end{tcolorbox}
\caption{Instruction prompt used by Omni-Judge for the audio-video-text semantic alignment metric.}
\label{prompt:audio_video_text_alignment}
\end{figure*}

\begin{figure*}[t]
\centering
\begin{tcolorbox}[colback=black!5!white,colframe=black!75!black,title=Audio-Video Temporal Alignment Prompt]
Job Description:\\
You are an evaluation assistant to generate the Audio-Video Temporal Alignment metric score. Your task is to evaluate the TIMING alignment between AUDIO and VIDEO.\\
Focus on whether sound events occur exactly when their visual counterparts happen.\\

Definition:\\
Temporal Synchronization measures the timing accuracy between visual and sound events.\\
It is about *when* the sound happens, not *what* it is.\\
Examples: lip movement matching speech timing, explosion sound matching flash timing, etc.\\

Inputs:\\
- video\_frames: generated video frames (with timestamps)\\
- audio\_track: generated audio\\

Output:\\
- One integer score from 1 to 5 indicating synchronization accuracy\\

Scoring Criteria:\\
1 (Very Poor): Large, persistent delays; sounds occur far too early or late.\\
2 (Poor): Frequent noticeable mismatches.\\
3 (Neutral): Moderate timing drift; generally acceptable.\\
4 (Good): Mostly synchronized; minor misalignments.\\
5 (Very Good): Perfectly timed; sound and motion align seamlessly.\\

Instruction:\\
Focus on the timing of audio and visuals. Judge whether sounds occur exactly when expected.\\
Output one integer score (1–5).\\
\end{tcolorbox}
\caption{Instruction prompt used by Omni-Judge for the audio-video temporal alignment metric.}
\label{prompt:audio_video_temporal_alignment}
\end{figure*}

\begin{figure*}[t]
\centering
\begin{tcolorbox}[colback=black!5!white,colframe=black!75!black,title=Audio Aesthetic Prompt]
Job Description:\\
You are an evaluation assistant to generate the Audio Aesthetic metric score. Your task is to rate the AESTHETIC quality of the AUDIO.\\
Focus on its artistic expression and emotional appeal.\\

Definition:\\
Audio Aesthetic evaluates whether the sound is pleasant, expressive, stylistically fitting, and emotionally engaging.\\
It reflects creativity and production taste, not technical clarity.\\

Inputs:\\
- audio\_track: generated audio\\
- text\_prompt: optional context for intended style or emotion\\

Output:\\
- One integer score from 1 to 5 indicating audio aesthetic quality\\

Scoring Criteria:\\
1 (Very Poor): Harsh or unpleasant; lacks structure or emotional intent.\\
2 (Poor): Unrefined, monotonous, or awkwardly mixed; weak expressiveness.\\
3 (Neutral): Average sound; lacks distinctive style or emotion.\\
4 (Good): Pleasing, expressive, and stylistically coherent.\\
5 (Very Good): Beautiful, emotionally rich, and artistically impressive.\\

Instruction:\\
Listen to the audio holistically. Rate its artistic and emotional appeal with a single integer (1–5).
\end{tcolorbox}
\caption{Instruction prompt used by Omni-Judge for the audio aesthetic metric.}
\label{prompt:audio_aesthetic}
\end{figure*}

\begin{figure*}[t]
\centering
\begin{tcolorbox}[colback=black!5!white,colframe=black!75!black,title=Video Aesthetic Prompt]
Job Description:\\
You are an evaluation assistant to generate the Video Aesthetic metric score. Your task is to evaluate the artistic and stylistic appeal of the VIDEO.\\
Focus on composition, lighting, color grading, and cinematic presentation.\\

Definition:\\
Video Aesthetic measures the visual artistry and stylistic coherence of a video — how beautiful, well-composed, and cinematic it looks.\\
It captures aspects such as framing, camera movement, lighting design, color harmony, and overall visual tone, independent of the video’s realism or semantic content.\\

Preferred characteristics:\\
- Film-like visuals with balanced lighting, clear focus, and consistent tone.\\
- Bright, sharp, high-resolution imagery with vivid yet natural color grading.\\
- A cohesive and pleasing visual style across frames.\\

Unfavorable characteristics:\\
- Dull, overly dark, or muddy visuals.\\
- Ugly monster or creature.\\
- Blood or violation.\\
- Unclear composition, awkward camera angles, or distorted framing.\\
- Strange or unpleasant color palettes or stylistic choices that reduce aesthetic appeal.\\

Inputs:\\
- video\_frames: generated video frames\\

Output:\\
- One integer score from 1 to 5 representing visual aesthetic quality\\

Scoring Criteria:\\
1 (Very Poor): Visually unpleasant; chaotic composition, poor lighting or color; dark or distorted appearance.\\
2 (Poor): Weak sense of style; unbalanced lighting, awkward framing, or dull tone.\\
3 (Fair/Neutral): Acceptable but plain or inconsistent style; lacks clear aesthetic intent.\\
4 (Good): Visually pleasing and coherent; good balance of light, color, and composition.\\
5 (Excellent): Film-like, bright, sharp, and beautifully composed visuals with strong artistic polish.\\

Instruction:\\
Watch the video frames carefully and judge their overall beauty, clarity, and cinematic style.\\
Prefer visually bright, balanced, and aesthetically refined videos.\\
Output only one integer score (1–5).\\
\end{tcolorbox}
\caption{Instruction prompt used by Omni-Judge for the video aesthetic metric.}
\label{prompt:video_aesthetic}
\end{figure*}

\end{document}